\documentclass[runningheads]{llncs}
\usepackage{graphicx}
\usepackage{multicol}
\usepackage{multirow}
\usepackage{tikz}
\usepackage{amsmath}
\usepackage{amsfonts}
\usepackage{arydshln} 

\usepackage[inline]{enumitem}

\begin{document}
%
\title{BMFT: Achieving Fairness via Bias-based Weight Masking Fine-tuning}
%
\titlerunning{BMFT: Achieving Fairness via Bias-based Weight Masking Fine-tuning}
%
\author{Yuyang Xue\inst{1}\thanks{The two authors share the equal contribution.} \and 
Junyu Yan\inst{1}$^\star$ \and
Raman Dutt\inst{1, 2} \and
Fasih Haider\inst{1} \and
Jingshuai Liu\inst{1} \and
Steven McDonagh\inst{1} \and 
Sotirios A. Tsaftaris\inst{1}}
\authorrunning{Y. Xue et al.}
%
\institute{School of Engineering, The University of Edinburgh, Edinburgh, EH9 3FG, UK \email{\{yuyang.xue,junyu.yan,fasih.haider,jliu11,s.mcdonagh,s.tsaftaris\}@ed.ac.uk} \and
School of Informatics, The University of Edinburgh, Edinburgh, EH8 9AB, UK
\email{\{r.dutt\}@sms.ed.ac.uk}}
%
\maketitle              
\begin{abstract}
Developing models with robust group fairness properties is paramount, particularly in ethically sensitive domains such as medical diagnosis. Recent approaches to achieving fairness in machine learning require a substantial amount of training data and depend on model retraining, which may not be practical in real-world scenarios. 
To mitigate these challenges, we propose Bias-based Weight Masking Fine-Tuning (BMFT), a novel \textit{post-processing} method that enhances the fairness of a trained model in significantly fewer epochs without requiring access to the original training data. BMFT produces a mask over model parameters, which efficiently identifies the weights contributing the most towards biased predictions. Furthermore, we propose a two-step debiasing strategy, wherein the feature extractor undergoes initial fine-tuning on the identified bias-influenced weights, succeeded by a fine-tuning phase on a reinitialised classification layer to uphold discriminative performance.
Extensive experiments across four dermatological datasets and two sensitive attributes demonstrate that BMFT outperforms existing state-of-the-art (SOTA) techniques in both diagnostic accuracy and fairness metrics. Our findings underscore the efficacy and robustness of BMFT in advancing fairness across various out-of-distribution (OOD) settings. Our code is available at: \url{https://github.com/vios-s/BMFT}


\end{abstract}
\section{Introduction}

Machine learning is known to exhibit biases from various sources, such as human judgement, inherent algorithmic predispositions, and representation bias~\cite{mehrabi2019survey}. 
The latter emerges when data from minority ethnicities, genders, and age groups are under-represented, leading to unfair predictions by machine learning systems. This can result in serious issues, including direct or indirect discrimination~\cite{pagano2023bias}. 
It is important to ensure an equitable and responsible application of AI to not undermine the trustworthiness and acceptance of AI solutions by end users~\cite{winkler2019association}.


Xu et al.~\cite{xu2023fairness} identified three key strategies to mitigate bias in medical imaging: pre-processing, 
e.g.~sample re-distribution~\cite{puyol2021fairness}; in-processing, including learning of disentangled representations~\cite{bissoto2020debiasing,liu2021disentangled}; and \textit{post-processing}, involving weight pruning~\cite{marcinkevics2022debiasing,wu2022fairprune}.
While pre- and in-processing are effective in producing fair predictions, a challenge remains: \textit{Access to original medical training data, post-deployment in the real-world, is often restricted}.
Mao et al.~\cite{mao2023last} showed that empirical risk minimisation (ERM) on imbalanced data can effectively capture features for classification within the representation domain. 
This suggests that post-processing can overcome bias with only a small amount of external data, {e.g.}~by fine-tuning or pruning weights, thus eliminating the need for complete model retraining on the large-scale original training dataset. 

Weights learnt with ERM encode two types of features: \emph{core} features that contribute to the target task and \emph{bias} features that have captured spurious correlations or irrelevant information in the data and could therefore exhibit bias during decision-making. It is well understood that these two distinct types of features are entangled in a complex manner~\cite{le2023last,ye2023freeze}.  
Current weight pruning strategies~\cite{wu2022fairprune,huang2023mitigating} aim to identify and remove weights that contribute to bias, based on certain heuristics. However, this can cause substantial performance degradation mainly due to the complex entanglement between \textit{core} features and bias features, especially when a model is trained from random initialisation~\cite{tran2022pruning}. 
It is well known that fine-tuning from a pre-trained initialisation leads to faster convergence and better performance~\cite{raghu2019transfusion}. However, if fine-tuning is indiscriminately applied to all weights in pursuit of fairness, there is the potential of updating \textit{core} features which have previously captured essential discriminative information, therefore risking predictive performance power, even if model fairness gets improved.
Moreover, fine-tuning all model parameters with limited data is prone to overfitting~\cite{dutt2023parameter}.
A potential solution is to fine-tune a subset of model weights to lessen the effect of biased feature on the decision boundary~\cite{le2023last}.

We propose Bias-based Weight Masking Fine-Tuning (BMFT), a fast, retrain-free, post-processing debiasing approach designed to enhance fairness while preserving performance, which utilises a small external dataset and eliminates the need for original training data. Different from other mask-based methods that implement iterative re-training on the original data~\cite{dutt2023fairtune}, \textit{BMFT} firstly generates a parameter update mask which highlights the model parameters contributing to bias. Then, a two-step process fine-tunes the feature extractor on mask-selected weights for feature debiasing, which is followed by classification layer fine-tuning to efficiently integrate core features and improve classification performance. This strategy can lead to remarkable performance improvements in terms of prediction fairness and accuracy within significantly fewer training steps.
Our \textbf{contributions} can be summarised as follows:
\textbf{(1)} We propose a fast mask generation technique to filter weights contributing to bias with a small external dataset and without needing the original training data.
\textbf{(2)} We introduce a two-step fine-tuning strategy that is model-agnostic and contributes to bias mitigation without sacrificing the model's performance.
\textbf{(3)} We conduct extensive experiments across different datasets and sensitive attributes to reveal the superiority of our proposed approach compared to SOTAs.

\section{Method}

\begin{figure}[t]
    \centering
    \includegraphics[width=\textwidth]{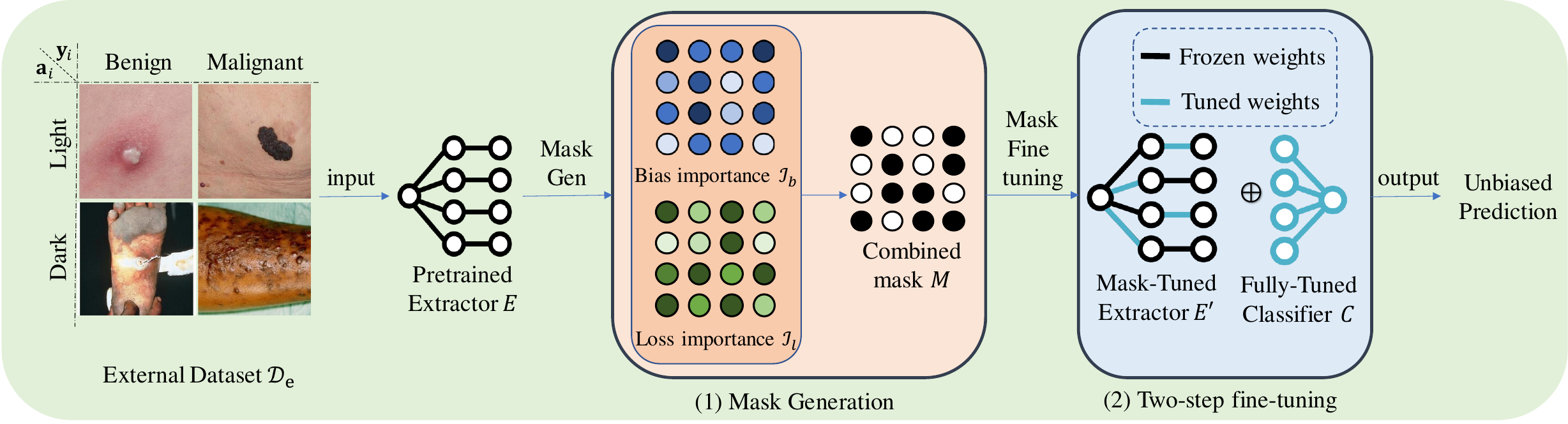}
    \caption{\textit{BMFT} is a masked-based fine-tuning post-processing approach. (1) A weight mask is generated by calculating weight importance to the bias and loss functions. (2) The masked weights of the feature extractor $E(\cdot)$ are fine-tuned to debias, followed by fine-tuning a reinitialised classification layer $C(\cdot)$ to maintain model performance.}\label{fig1}
\end{figure}

Bias in model prediction typically originates from two key sources: (1) the \textit{entanglement} between noisy, harmful (spurious or irrelevant) features and useful, \emph{core} features within the feature extractor; and (2) the \textit{incorrect composition} of representation in the classification layer, leading to the core features information loss~\cite{mao2023last}.
\textbf{The main idea} of \textit{BMFT} Fig. \ref{fig1} is to start with debiasing the feature extractor, by fine-tuning only the targeted weights that are disproportionately instigating bias. 
Subsequently, as the first process may affect how the core features are integrated by the classification layer for prediction, \textit{BMFT} then fine-tunes a reinitialised classification layer. 

\subsection{Preliminaries} 

Consider an external dataset of $N$ subjects,
$\mathcal{D}_\mathrm{e}=\{\mathbf{x}_{i}, \mathbf{y}_{i}, \mathbf{a}_{i}\}^N_{i=1}$, where $\mathbf{x}_{i}$, $\mathbf{y}_{i}$, and $\mathbf{a}_{i}$ are the input image, the corresponding label, and the sensitive attribute, respectively. 
For simplicity, we assume binary targets and attributes (i.e.~$\mathbf{y}_i, \mathbf{a}_{i} \in \{0, 1\}$). 
We will assume a compositional construction of a biased model $f_{\theta}(\cdot)$, consisting of a feature extractor $E(\cdot)$ and a classification head $C(\cdot)$, pre-trained by the original training data $\mathcal{D}_\mathrm{org}$. 
Our goal is to debias the model $f_{\theta}(\cdot)$ w.r.t.~fairness metrics such as Demographic Parity (DP)~\cite{dwork2012fairness} or Equalised Odds (EOdds)~\cite{hardt2016equality} by fine-tuning with significantly fewer epochs, using $\mathcal{D}_{\mathrm{e}}$; and further enhance the generalisability of the model on various OOD test datasets $\mathcal{D}_\mathrm{t}$, which share the same sensitive attribute $\mathbf{a}_{i}$. 
To avoid attribute bias from the \textit{external dataset} $\mathcal{D}_{\mathrm{e}}$, we sample a \textit{group-balanced subset} $\mathcal{D}_{r}$ from $\mathcal{D}_{\mathrm{e}}$. This ensures that each attribute group $\mathbf{a}_{i}$ has the same amount of data. 
The proportion of positive to negative labelled images is adjusted to be the same in each $\mathbf{a}_{i}$ group, which prevents bias due to different label distributions. 
Implementing these two strategies prepares the model for debiasing by providing equal representation across attribute groups and equal ratios of label types. Choosing proportionally over numerical equality in $\mathcal{D}_{r}$ helps when there are few positive or negative labels in $\mathcal{D}_{\mathrm{e}}$, thus avoiding a reduction in the overall sample size.

\subsection{Bias-Importance-Based Mask Generation}

To identify the effective parameters for updating, we evaluated the importance of each parameter with respect to the designed bias influence function and loss function. 
The top $K$ 
parameters, that are highly correlated with bias and yet provide only minimal contribution to prediction, were selected as the masking weights.
We adopt weighted Binary Cross Entropy (WBCE) as loss function since the dermoscopic dataset contains label imbalance. 
Together with the bias influence function $\mathcal{B}$, the differentiable
proxy of EOdds~\cite{hardt2016equality},
is given by:
\begin{equation}
   \mathcal{L}_{\mathrm{WBCE}}=\sum_i \left(-\frac{N_n}{N_n+N_p}\mathbf{y}_i\log(p_i)-\frac{N_p}{N_n+N_p}(1-\mathbf{y}_{i})\log(1-p_i)\right),
\label{bce_loss}
\end{equation}
\begin{equation}
     \mathcal{B}(f_{\theta}, \mathbf{x}_{i}, \mathbf{y}_{i}, \mathbf{a}_{i}) = tpr + fpr,
\label{eq:bias_EOdds}
\end{equation}
where 
\begingroup
\small
\begin{equation}
\begin{aligned}
    tpr &= \left|
    \frac{\sum_{i}(1\!-\!a_{i})y_{i}f_{\theta}(x_{i})}{\sum_{i}(1\!-\!a_{i})y_{i}} \!-\! \frac{\sum_{i}a_{i}y_{i}f_{\theta}(x_{i})}{\sum_{i}a_{i}y_{i}}\right|\raisebox{-2ex}{,}\ \\
    fpr &= \left|
    \frac{\sum_{i}\!(1\!-\!a_{i})(1\!-\!y_{i})f_{\theta}(x_{i})}{\sum_{i}(1\!-\!a_{i})(1\!-\!y_{i})}\!-\! \frac{\sum_{i}a_{i}(1\!-\!y_{i})f_{\theta}(x_{i})}{\sum_{i}a_{i}(1\!-\!y_{i})}\right| \nonumber
\end{aligned}\raisebox{-5ex}{.}\
\label{bias_eodds}
\end{equation}
\endgroup

where $p_i$ is the prediction probability, $N_{p}$ is the number of images with positive label, $N_{n}$ the number of images with negative label. For a pre-trained model $f_{\theta}(\cdot)$ with effectively learnt parameters $\theta^{*}$, the sensitivity for each parameter $\theta$ can be obtained by the second-order derivative of loss close to the minimum~\cite{foster2023fast}.
This value can be interpreted as the importance of the parameter for the prediction. 
The diagonal of the Fisher Information Matrix (FIM) is equivalent to the second-order derivative of the loss~\cite{pawitan2001all}, which provides an efficient computation through first-order derivatives. 
The first-order expression for the FIM is provided by:
\begin{equation}
    \mathcal{I}(f_\theta, \mathcal{D}_{\mathrm{e}}) = \mathbb{E}\left[\left(\frac{\partial \log f_\theta(\mathcal{D}_{\mathrm{e}}|\theta)}{\partial\theta}\right)\left(\frac{\partial \log f_\theta(\mathcal{D}_{\mathrm{e}}|\theta)}{\partial\theta}\right)^\top\bigg|_{\theta}\right]
    \label{FIM}.
\end{equation}

We employ Eq.~\ref{FIM} to quantify the importance of each weight that contributes to the prediction. The importance of each weight for the bias can be computed using the second derivative of the bias influence function (that is, Eq.~\ref{bias_eodds}).
Noting that the bias influence function is a linear function of the prediction, and thus replaces the loss function with the bias influence function during backpropagation, the importance of a parameter w.r.t.~the bias can also be derived by Eq.~\ref{FIM}. We denote the weight mask as $M_i$, where $i$ is the weight index.
%


To ensure that the weight importance for bias or prediction is not skewed in favour of a particular subgroup, in practice, we use the group-balanced dataset $\mathcal{D}_{\mathrm{r}}$ to calculate the importance. 
To select top $K\%$ weights in the mask $M_i$, we use the bias importance $\mathcal{I}_{i, b}$ and the loss importance $\mathcal{I}_{i, l}$ of each parameter as:
\begin{equation}
    M_{i} = \left\{
\begin{array}{rcl}
1, \quad     &      & \mathrm{if} \ \ \mathcal{I}_{i,b}/ \mathcal{I}_{i, l} > \alpha \\
0, \quad     &      & \text{otherwise}\\
\end{array} \right.
\label{mask},
\end{equation}
where $\mathcal{I}_{i,b}$ and $\mathcal{I}_{i,l}$ denote the importance of each parameter in relation to bias and loss, respectively. The hyperparameter $\alpha$ distinguishes the significance of the weight to bias-induced features as opposed to core features, calculated by using the weight selection ratio $K$, defined as the top $K$-th percentile of the ratios of importance of the parameters. 
An illustrative example of a mask is provided in the supplementary material.

\subsection{Impair-Repair: Two-Step Fine-Tuning}

To disentangle core features from biased features, whilst preserving predictive capability, a two-stage fine-tuning process is structured in an ``impair-repair'' manner.
Firstly, we fine-tune the selected masked parameters within the feature extractor $E(\cdot)$ using the group-balanced dataset $\mathcal{D}_{\mathrm{r}}$, guided by an objective function (Eq.~\ref{bias_eodds}). Next, we fine-tune the entire reinitialised classification layer $C(\cdot)$.
The weights identified by the mask in the feature extractor are closely linked to attribute bias and play a smaller role in predicting the class. This relationship helps us remove the unfavourable influence of bias contained in the model, and yet crucially also retain information embedded in core features.

We highlight that the pre-trained model has the capability to sufficiently extract core features. Further, eliminating the influence of (only) bias-effected weights suggests moving the model decision boundary away from bias features. Moreover, fine-tuning is prone to overfitting on a small dataset. These observations suggest that small epochs are sufficient to realise the proposed debiasing fine-tuning strategy. 
%

An unbiased model, trained with a balanced dataset, will have the majority of its classification layer weights being zero, due to minimisation of irrelevant feature impact on class prediction~\cite{le2023last}. 
To this end, we first reinitialise the weights and biases in the classification layer $C(\cdot)$ to zero to enable fast convergence to the optimal unbiased state and then fine-tune the layer on $\mathcal{D}_{\mathrm{r}}$ using a loss which combines the WBCE loss $\mathcal{L}_{\mathrm{WBCE}}$ and fairness constraints (i.e.~Eq.~\ref{bias_eodds}), as:
\begin{equation}
    \mathcal{L}=\mathcal{L}_{\mathrm{WBCE}}(f_{\theta}) + \beta \mathcal{B}(f_{\theta}),
    \label{repairobj}
\end{equation}
where the fairness constraint $\mathcal{B}(f_{\theta})$ modulates the emphasis of the features, and the hyperparameter $\beta$ establishes an efficiently identified core trade-off between class prediction accuracy and fairness.  
The improved feature extractor retains core features without being influenced by attribute bias, allowing the model to efficiently identify and recover its predictive performance in limited epochs. 
To balance bias reduction with efficiency, an equal number of epochs is used for both the ``impair'' and ``repair'' stages. 
Our empirical findings indicate that only 10\% of the total original training epochs are adequate for successful fine-tuning.

\section{Experiments and Results}

\subsection{Dataset and Data-processing}

\noindent \textbf{ISIC Challenge Training Dataset.} 
The International Skin Imaging Collaboration (ISIC) challenge dataset
is a collection of dermoscopic images of melanoma classification, complete with diagnostic labels and metadata. 
A combination of the 2017~\cite{codella2018skin}, 2019~\cite{tschandl2018ham10000,combalia2019bcn20000} and 2020~\cite{rotemberg2021patient} ISIC challenge data is used as the training dataset (46,938 images), while the 2018~\cite{codella2019skin} data is reserved as the \textit{external dataset} $\mathcal{D}_e$ (9,925 images, excluding any overlap with the training data).
We select 4,024 images for skin tone and 9,084 for gender from  $\mathcal{D}_e$ to build \textit{group-balanced subset} $\mathcal{D}_r$. 
We maintain the same skin tone annotation as~\cite{bevanskin} for training data.
All images are pre-processed with centre-cropping and resizing to size $256{\times}256$. We consider two available sensitive attributes (skin tone and gender).

\noindent \textbf{Test Datasets.} 
Our study employs four different OOD datasets with skin images for melanoma detection.
Fitzpatrick-17k~\cite{groh2021evaluating} contains 16,577 images across six skin tone levels, which we group into light (1-3) and dark (4-6) categories.  
The PAD-UFES-20~\cite{pacheco2020pad} includes 2,298 images, with detailed metadata such as gender, ethnicity, and skin tone. 
Interactive Atlas of Dermoscopy (Atlas)~\cite{lio2004interactive} provides 1,011 images with gender, and DDI~\cite{daneshjou2022disparities} offers 656 images with age and skin information. 
These datasets expand diversity, aiding in evaluating our model's generalisability across attributions. 
Details are in the supplementary material.

\subsection{Implementation}

We conducted experiments in PyTorch~\cite{paszke2019pytorch} using NVIDIA A100 40GB GPUs.
We adopt different variants of ImageNet pretrained ResNet architecture (ResNet-18, 34, 50, and 101)~\cite{he2016deep}. The learning rates for impair and repair processes were set at 0.001 and 0.003 respectively, using Adam optimiser~\cite{kingma2014adam}, with a batch size of 64. We perform a hyperparameters sweep and selected 
$K=50$, and $\beta$ was selected as 0.02 for the ResNet50 model (details on hyperparameter tuning are shown in supplementary materials). 
To address class-imbalance, we adopt the WBCE loss function $\mathcal{L}_{\mathrm{WBCE}}$. Furthermore, we use accuracy (ACC), area under the receiver operating characteristic curve (AUC) as primary performance metrics, and equalised odds (EOdds~\cite{hardt2016equality}) as fairness metric, as done previously~\cite{dutt2023fairtune,han2018classification,wu2022fairprune}.
Our post fine-tuning method requires just an additional five epochs, which is 10\% of the initial 50-epoch training duration. 
Code will be released soon. 


\subsection{Results}

We compare with baselines and SOTA models which we describe below. 
We pre-trained the ResNet model on the training dataset $\mathcal{D}_{\mathrm{org}}$ as a \textit{Baseline}. Performance is evaluated using established metrics. 
Full fine-tuning (\textit{Full FT}), a widely adopted fine-tuning practice, adjusts all weights on the external dataset $\mathcal{D}_{\mathrm{e}}$ with the loss function specified in Eq.~\ref{repairobj}. 
\textit{FairPrune}~\cite{wu2022fairprune} improves fairness by pruning parameters and prioritises subgroup accuracy; however, the optimal hyperparameter configuration is specific to model and dataset.
\textit{LLFT}~\cite{mao2023last} fine-tunes only the last layer of a deep classification model to promote fairness.
Similarly, \textit{Diff-Bias}~\cite{marcinkevics2022debiasing}  fine-tunes on an external dataset, using a bias-aware loss function to steer network optimisation. 
We highlight that a good fairness score 
does not constitute the sole criterion for success and that a model should also afford accurate classification behaviour, 
represented here by high ACC and AUC. 
The results presented here correspond to the ResNet-50 model and analogous trends were observed across the investigated models, with further details present in the supplementary materials.

\begin{table}[t]
\caption{Comparison of debiasing methods on skin tone and gender attributes, for a ResNet-50 model. We re-implemented SOTA methods (denoted with $^*$) or used publicly available code (\textsuperscript{\textdagger}).
\textit{FairPrune} had difficulty in classifying the Atlas dataset.}\label{tab2}
\centering
\resizebox{\textwidth}{!}{
\begin{tabular}{l|ccc:ccc:ccc|ccc:ccc}
 & \multicolumn{9}{c|}{\textbf{Skin Tone}} & \multicolumn{6}{c}{\textbf{Gender}}\\\hline
\multirow{2}{*}{\textbf{Methods}}& \multicolumn{3}{c:}{\textbf{Fizpatric17k}} & \multicolumn{3}{c:}{\textbf{PAD-UFES-20}} & \multicolumn{3}{c|}{\textbf{DDI}} & \multicolumn{3}{c:}{\textbf{Atlas}} & \multicolumn{3}{c}{\textbf{PAD-UFES-20}} \\
& \multicolumn{1}{|c}{ACC↑} & \multicolumn{1}{c}{AUC↑} & \multicolumn{1}{c:}{EOdds↓} & \multicolumn{1}{c}{ACC↑} & \multicolumn{1}{c}{AUC↑} & \multicolumn{1}{c:}{EOdds↓} & \multicolumn{1}{c}{ACC↑} & \multicolumn{1}{c}{AUC↑} & \multicolumn{1}{c|}{EOdds↓} & \multicolumn{1}{c}{ACC↑} & \multicolumn{1}{c}{AUC↑} & \multicolumn{1}{c:}{EOdds↓} & \multicolumn{1}{c}{ACC↑} & \multicolumn{1}{c}{AUC↑} & \multicolumn{1}{c}{EOdds↓} \\\hline
{Baseline}  & 0.719 & 0.521 & 0.1745 & 0.858 & 0.506 & 0.0412 & 0.744 & 0.646 & 0.1288 & 0.574 & 0.614 & 0.0348 & 0.858 & 0.506 & 0.0167\\
{Full FT}   & 0.703 & 0.522 & 0.1510  & 0.725 & 0.501 & \textbf{0.0321}  & 0.727 & 0.618 & 0.2059& 0.690 & 0.655 & 0.0303 & 0.725 & 0.501 & 0.0730 \\
{FairPrune~\cite{wu2022fairprune}$^*$} & 0.763 & 0.547 & 0.0120 & 0.698 & 0.492 & 0.1237 & 0.737 & 0.536 & \textbf{0.0024}& $\backslash$ & $\backslash$ & $\backslash$ & 0.634 & 0.499 & 0.1456 \\
{LLFT~\cite{mao2023last}\textsuperscript{\textdagger}}      & 0.579 & 0.513 & 0.1446 & 0.694 & 0.492 & 0.0726 & 0.646 & 0.561 & 0.1197& 0.724 & 0.589 & 0.0042& \textbf{0.964} & 0.482 & \textbf{0.0007}\\
{Diff-Bias~\cite{marcinkevics2022debiasing}\textsuperscript{\textdagger}} & 0.716 & 0.507 & 0.1144 & 0.884 & 0.501 & 0.0599 & 0.737 & 0.629 & 0.0741 & \textbf{0.769} & 0.687 & 0.0383 & 0.878 & 0.504 & 0.0461 \\\hline
{BMFT (Ours)}      & \textbf{0.865} & \textbf{0.637} & \textbf{0.0055} &\textbf{0.902} &\textbf{0.507} &0.0657 &\textbf{0.759} &\textbf{0.736} &0.0754 & 0.752 & \textbf{0.876} & \textbf{0.0037} & 0.893 & \textbf{0.512} & 0.0365   \\\hline
\end{tabular}}
\end{table}


\noindent \textbf{Fine-tuning can help achieve fairness.}
We start by evaluating the behaviour of fine-tuning methods. 
In terms of fairness, we observed significant improvements both in AUC and EOdds across the majority of test datasets when employing fine-tuning-based methods (Table~\ref{tab2}). This indicates that fine-tuning can improve the accuracy gaps between two attribute groups and improve fairness metrics. 
Fine-tuning with fairness constraints, i.e.~\textit{LLFT} and \textit{Diff-Bias}, exhibit lower EOdds compared to using a WBCE loss, i.e.~\textit{Full FT}. This illustrates that incorporating the bias influence function into the loss has measurable benefits. 

\noindent \textbf{Pruning is not always the best option.}
Model pruning, a popular technique for model simplification towards reducing bias, does not always lead to fairer outcomes, c.f.~fine-tuning methods. 
Our results demonstrate that pruning achieves low AUC across most test datasets, even under-performing the baseline. For example, the performance of \textit{FairPrune} drops drastically under the Atlas dataset, as detailed in Table~\ref{tab2}, regardless of hyperparameter values. 
It is ``easier'' to obtain good fairness by finding less useful features for classification. This leads to poor classification results in sensitive groups.
We conjecture that such performance declines are due to pruning which removes neurons containing \textit{both} bias and core features that are important for classification, therefore causing undesirable information loss. 
Balancing between fairness and classification performance through pruning requires extensive iterations and is heavily dependent on hyperparameters and datasets. 
Instead, our method considers both classification capability and fairness to force the selection of core features.


\noindent \textbf{Masked fine-tuning utility.}
Traditional fine-tuning techniques struggle to achieve a balance between classification performance and fairness. 
\textit{Full FT} and \textit{Diff-Bias} improve all metrics only in the Atlas dataset for the gender attribute (see Table~\ref{tab2}). This illustrates that fine-tuning all layers guides model behaviour towards the distribution of the external dataset, therefore necessitating similar distributions between external and test datasets. Moreover, this may affect weights that are useful for prediction in lieu of 
fairness, leading to degradation in classification performance.
\textit{LLFT} shows improvement in EOdds, but performs poorly in terms of AUC (0.589 in Atlas), showing that fine-tuning the last layer from scratch requires all core features that are well captured and isolated from the bias features; otherwise, 
fairness and prediction irrelevant features 
may be selected by the classification layer for achieving fairness. 
The proposed method, \textit{BMFT}, manages to find an ideal balance and exhibits the best performance in terms of ACC, AUC and EOdds on most test datasets. 

\noindent \textbf{Different bias, different difficulty.} Comparing \textit{Baseline} results for gender and skin tone attributes in the PAD-UFES-20 dataset (see Table~\ref{tab2}), the pre-trained model demonstrates less inherent bias for gender, indicated by a lower EOdds. Furthermore, Table~\ref{tab2} shows that \textit{BMFT} achieves more prominent EOdds and AUC improvements for the gender attribute c.f.~the skin tone. This finding suggests that attributes with less inherent bias in the pre-trained model are easier for bias mitigation strategies to rectify. Moreover, the presence of less inherent bias suggests that core and biased features are less entangled, 
which simplifies the task of maintaining model performance during mitigation process.

\noindent \textbf{Mask: better than manual, superior to random.}
\begin{table}[t]
\caption{Comparison of the random mask, manual masks, and our proposed mask generation method on both skin and gender attributes, using ResNet-50 backbone.}\label{tab4}
\centering
\resizebox{0.7\textwidth}{!}{
\begin{tabular}{l|ccc|ccc}
\multirow{2}{*}{Methods}  & \multicolumn{3}{c|}{DDI (Skin Tone)} & \multicolumn{3}{c}{Atlas (Gender)} \\
& \multicolumn{1}{|c}{ACC↑} & \multicolumn{1}{c}{AUC↑} & \multicolumn{1}{c}{EOdds↓} & \multicolumn{1}{|c}{ACC↑} & \multicolumn{1}{c}{AUC↑} & \multicolumn{1}{c}{EOdds↓} \\\hline
{Random}  &0.733	&0.581	&0.1194 & 0.741	& 0.629	&0.0105\\
{Manual Top Layers}   & 0.749&0.670 &0.0948 &0.753 & 0.644&0.0169 \\
{Manual BN Layers} & 0.741 & 0.635 & 0.1116 &\textbf{0.757} &0.661 &0.0052 \\\hline
{BMFT (Ours)}      &\textbf{0.759} &\textbf{0.736} &\textbf{0.0754}& 0.752 & \textbf{0.876} & \textbf{0.0037}\\
\end{tabular}}
\end{table}
We compare our mask with random masking, manual shallow convolutional (Top) masking and batch normalisation (BN) masking in Table~\ref{tab4}. The latter two options are motivated by analysis of the mask distribution, where we saw that mask-selected weights exist mainly in shallow convolutional layers and BN. 
Compared to random, the results align with our hypothesis that shallow convolutional layers and BN layers contribute to spurious correlation. 
Our method outperforms the baselines by ${\sim}0.2$ in AUC, illustrating the effectiveness of our fast mask generation process. 
With masking, spurious-related weights can be fine-tuned, and core-feature-related weights will remain intact, leading to improved fairness and prediction efficacy. \\

\section{Conclusion}
Our study advances bias mitigation in discriminative models for dermatological disease. The proposed \textit{BMFT} distinguishes the core features from biased features, enhancing fairness without sacrificing classification performance. This two-step impair-repair fine-tuning approach can effectively reduce bias within minimal epochs, with only 10\% of the original training effort, providing an efficient solution when computing resources or data access are limited. 
Our method is agnostic to the choice of pre-trained models.
Our future work will explore a broader range of tasks and multiple attribute-label pairs. The challenge of obtaining diverse, well-labelled public datasets with comprehensive meta-data or sensitive attributes remains a limitation on our work.

\begin{credits}
\subsubsection{\ackname}Y. Xue and J. Yan thank additional financial support from the School of Engineering, the University of Edinburgh. J.Yan also thanks the support of Advanced Care Research Center. S.A. Tsaftaris also acknowledges the support of Canon Medical and the Royal Academy of Engineering and the Research Chairs and Senior Research Fellowships scheme (grant RCSRF1819\textbackslash8\textbackslash25), of the UK’s Engineering and Physical Sciences Research Council (EPSRC) (grant EP/X017680/1) and the National Institutes of Health (NIH) grant 7R01HL148788-03.
We thank Dr. Edward Moroshko for his help and support.

\subsubsection{\discintname}
The authors have no competing interests to declare that are relevant to the content of this article.
\end{credits}

\newpage
\newpage

\bibliographystyle{splncs04} 
\bibliography{reference} 

\end{document}


\title{BMFT: Achieving Fairness via Bias-based Weight Masking Finetuning \\---\ Supplementary Material}

\authorrunning{Y. Xue et al.}
\author{Yuyang Xue \and 
Junyu Yan \and
Raman Dutt \and
Fasih Haider\and
Jingshuai Liu\and
Steven McDonagh\and 
Sotirios A. Tsaftaris}
\authorrunning{Y. Xue et al.}
%
\institute{School of Engineering, The University of Edinburgh, Edinburgh, EH9 3FG, UK} 
\maketitle
\titlerunning{BMFT: Achieving Fairness via Bias-based Weight Masking Finetuning}
%

\appendix


\begin{figure}[b]
    \centering
    \includegraphics[width=0.9\textwidth]{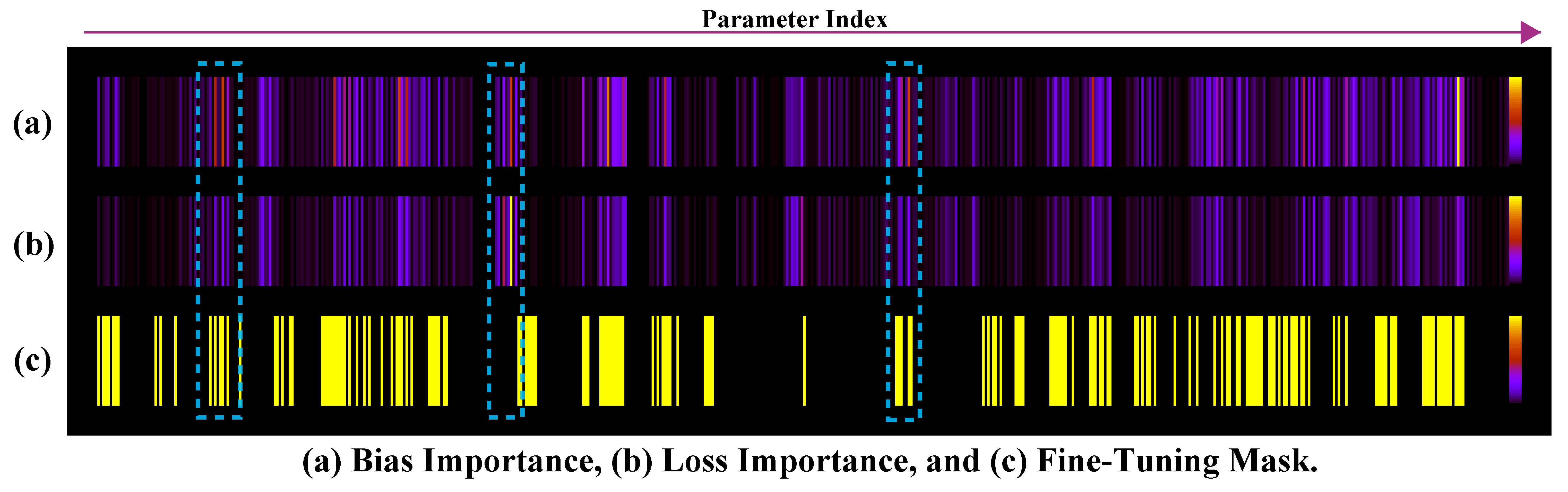}
    \caption{Mask generation starts by assessing each parameter for bias importance (a) and loss importance (b) (shown in heatmap, the higher the brighter). The ideal weights are those with high bias relevance and low loss impact. The final mask (c) is created using the mask, with blue dotted boxes in examples showing how these two are combined.}\label{sfig1}
\end{figure}

\begin{table}[]
\caption{The specifications for all datasets we used. Note that PAD-UFES-20 is extremely imbalanced in target. ``$\backslash$'' means the dataset does not contain this attribute.}\label{tab1}
\centering
\begin{adjustbox}{width=0.8\textwidth}
\begin{tabular}{l|cccccc}
Dataset       & Images & Benign & Malignant & Target & Skin Tone    & Gender           \\\hline\hline
ISIC-training & 46,938  & 43,142  & 3,796      & 1:11  & 6:1            & 1:1            \\
ISIC-external      & 9,925   & 8,818   & 1,107      & 1:8   & 4:1            & 1:1            \\\hline
Fitzpatrick   & 16,576  & 14,419  & 2,157      & 1:7   & 2:1            & \textbackslash{} \\
PAD-UFES-20   & 1,495   & 1,443   & 52        & 1:28  & 19:1           & 1:1            \\
DDI           & 657    & 486    & 171       & 1:3   & 2:1            & \textbackslash{} \\
Atlas         & 1,012   & 760    & 252       & 1:3   & \textbackslash{}& 1:1        \\\hline   
\end{tabular}
\end{adjustbox}
\end{table}
\vspace{-2em}

\begin{table}[]
\caption{Searching range for hyperparameters.}\label{tab1}
\centering
\begin{adjustbox}{width=0.7\textwidth}
\begin{tabular}{l|c}
Hyperparameters      &   Search Range\\\hline\hline
 impair learning rate & $[1\times10^{-2}, 1\times10^{-3}, 3\times10^{-3}, 5\times10^{-4}$] \\\hline
repair learning rate & $[1\times10^{-2}, 1\times10^{-3}, 3\times10^{-3}, 5\times10^{-4}$]\\\hline
$K$ & $[50, 70, 90, 95]$\\\hline
$\beta$ & $[0.01, 0.05, 0.02, 0.01]$\\\hline
\end{tabular}
\end{adjustbox}
\end{table}


\begin{sidewaystable}
\centering
\caption{Comparison of post-process bias-removal methods on skin tone and gender attributions. Note that ``$\backslash$'' means the method cannot classify the test data very well.}\label{stab1}
\resizebox{\textwidth}{!}{
\begin{tabular}{l|c|ccc:ccc:ccc|ccc:ccc}
& & \multicolumn{9}{c|}{Skin Tone} & \multicolumn{6}{c}{Gender}\\\hline\hline
\multirow{2}{*}{Methods}  & {\multirow{2}{*}{Architecture}} & \multicolumn{3}{c:}{Fizpatric17k} & \multicolumn{3}{c:}{PAD-UFES-20} & \multicolumn{3}{c|}{DDI} &  \multicolumn{3}{c:}{Atlas} & \multicolumn{3}{c}{PAD-UFES-20} \\
& & \multicolumn{1}{|c}{ACC↑} & \multicolumn{1}{c}{AUC↑} & \multicolumn{1}{c:}{Eodds↓} & \multicolumn{1}{c}{ACC↑} & \multicolumn{1}{c}{AUC↑} & \multicolumn{1}{c:}{Eodds↓} & \multicolumn{1}{c}{ACC↑} & \multicolumn{1}{c}{AUC↑} & \multicolumn{1}{c|}{Eodds↓}& \multicolumn{1}{c}{ACC↑} &\multicolumn{1}{c}{AUC↑} & \multicolumn{1}{c:}{Eodds↓} & \multicolumn{1}{c}{ACC↑} & \multicolumn{1}{c}{AUC↑} & \multicolumn{1}{c}{Eodds↓} \\\hline\hline
\multirow{4}{*}{Baseline}  & ResNet-18   & 0.715 & 0.554 & 0.1414 & 0.685 & 0.522 & 0.0404 & 0.713 & 0.598 & 0.1441    & 0.688 & 0.414 & 0.0263 & 0.685 & 0.522 & 0.0610\\  
                           & ResNet-34   & 0.709 & 0.541 & 0.1098 & 0.805 & 0.519 & 0.0467 & 0.715 & 0.579 & 0.0912    & 0.531 & 0.507 & 0.0288 & 0.805 & 0.519 & 0.0741\\  
                           & ResNet-50   & 0.719 & 0.521 & 0.1745 & 0.858 & 0.506 & 0.0412 & 0.744 & 0.646 & 0.1288    & 0.574 & 0.614 & 0.0348 & 0.858 & 0.506 & 0.0167\\  
                           & ResNet-101  & 0.717 & 0.520 & 0.1396 & 0.880 & 0.522 & 0.0543 & 0.698 & 0.569 & 0.1397& 0.729 & 0.662 & 0.0674 & 0.880 & 0.522 & 0.1027\\\hline
\multirow{4}{*}{Full FT}   & ResNet-18   & 0.816 & 0.566 & 0.1021  & 0.892 & 0.504 & 0.0572  & 0.755 & 0.741 & 0.0422  &0.751  &0.661  &0.0662  & 0.892 & 0.504 &0.0166  \\   
                           & ResNet-34   & 0.749 & 0.536 & 0.1284  & 0.935 & 0.482 & 0.0910  & 0.733 & 0.577 & 0.0247&0.770  &0.685  &0.0271  &0.935  &0.482  &0.0053  \\
                           & ResNet-50   & 0.703 & 0.522 & 0.1510  & 0.725 & 0.501 & 0.0321  & 0.727 & 0.618 & 0.2059 &0.690  &0.655  &0.0303  &0.725  &0.501  &0.0730  \\
                           & ResNet-101  & 0.737 & 0.541 & 0.1346  & 0.863 & 0.524 & 0.1239  & 0.726 & 0.614 & 0.1060  &0.682  & 0.667  & 0.0098 &0.863  &0.524  &0.0105  \\\hline          
\multirow{4}{*}{FairPrune} & ResNet-18   & 0.575 & 0.488 & 0.1322 & 0.932 & 0.510 & 0.0273 & 0.739 & 0.620 & 0.0056 & $\backslash$ & $\backslash$ & $\backslash$ & 0.860 & 0.515 & 0.0688 \\ 
                           & ResNet-34   & 0.762 & 0.468 & 0.0800 & 0.851 & 0.493 & 0.0365 & 0.743 & 0.705 & 0.0148& $\backslash$ & $\backslash$ & $\backslash$ & 0.312 & 0.500 & 0.0569 \\
                           & ResNet-50   & 0.763 & 0.547 & 0.0120 & 0.698 & 0.492 & 0.1237 & 0.737 & 0.536 & 0.0024& $\backslash$ & $\backslash$ & $\backslash$ & 0.634 & 0.499 & 0.1456 \\
                           & ResNet-101  & 0.688 & 0.518 & 0.2027 & 0.338 & 0.505 & 0.1027 & 0.727 & 0.476 & 0.0291& $\backslash$ & $\backslash$ & $\backslash$ & 0.246 & 0.489 & 0.1649 \\\hline
\multirow{4}{*}{LLFT}      & ResNet-18   & 0.708 & 0.545 & 0.1680 & 0.446 & 0.486 & 0.1946 & 0.699 & 0.583 & 0.1094& 0.749 & 0.375 & 0.0013 & 0.964 & 0.482 & 0.0007 \\
                           & ResNet-34   & 0.661 & 0.534 & 0.1072 & 0.763 & 0.501 & 0.0482 & 0.675 & 0.574 & 0.1713& 0.746 & 0.374 & 0.0025 & 0.964 & 0.482 & {0.0006} \\
                           & ResNet-50   & 0.579 & 0.513 & 0.1446 & 0.694 & 0.492 & 0.0726 & 0.646 & 0.561 & 0.1197& 0.724 & 0.389 & 0.0041 & {0.964} & 0.482 & 0.0007  \\
                           & ResNet-101  & 0.764 & 0.533 & 0.1199 & 0.733 & 0.505 & 0.0348 & 0.698 & 0.597 & 0.1633& 0.750 & 0.625 & 0.0030 & 0.951 & 0.482 & 0.0035 \\\hline
\multirow{4}{*}{Diff-Bias} & ResNet-18   & 0.830 & 0.559 & 0.0764 & 0.834 & 0.506 & 0.0251 & 0.740 & 0.870 & 0.0041& 0.768 & 0.683 & 0.0131 & 0.868 & 0.511 & 0.0657 \\
                           & ResNet-34   & 0.770 & 0.542 & 0.0897 & 0.919 & 0.502 & 0.0498 & 0.740 & 0.634 & 0.0393& 0.749 & 0.656 & 0.0354 & 0.918 & 0.495 & 0.0022 \\
                           & ResNet-50   & 0.716 & 0.507 & 0.1144 & 0.884 & 0.501 & 0.0599 & 0.737 & 0.629 & 0.0741& 0.769 & 0.687 & 0.0383 & 0.878 & 0.504 & 0.0461 \\
                           & ResNet-101  & 0.763 & 0.533 & 0.0958 & 0.892 & 0.523 & 0.0791 & 0.705 & 0.574 & 0.1627& 0.752 & 0.678 & 0.0223 & 0.887 & {0.531} & 0.0261 \\\hline
\multirow{4}{*}{BMFT (Ours)}      & ResNet-18   & 0.862 & 0.640 & 0.0121 & {0.944} &0.523&0.0364 &0.749 &{0.873} &{0.0040}     &0.756  &0.806  &0.0086  &0.862  &0.507  &0.0049  \\
                           & ResNet-34   & {0.865} & {0.705} & {0.0033} &0.924 &0.505 &{0.0241} &0.742 &0.655 &0.0333 &{0.778}  &0.729  &0.0020  &0.913  &0.512  &0.0468  \\
                           & ResNet-50   & {0.865} & 0.637 & 0.0055 &0.902 &0.507 &0.0657 &{0.759} &0.736 &0.0754 &0.752  &{0.876}  &0.0037  & 0.893  &0.512  &0.0365  \\
                           & ResNet-101  & 0.829 & 0.560 & 0.0666 &0.912 &{0.526} &0.0621 &0.742 &0.871 &0.0082 &0.755  &0.752  &0.0010  &0.878  &0.518  &0.0404  \\\hline
\end{tabular}
}\end{sidewaystable}